\documentclass{dynafront2025} 

\usepackage{csquotes}
\usepackage{amsmath,amssymb}
\usepackage{graphicx}
\usepackage{wrapfig}
\usepackage{tikz}
\usepackage{dirtytalk}
\usetikzlibrary{shapes.geometric, arrows.meta, positioning, calc, shadows, patterns, decorations.pathmorphing, backgrounds}

\title[Metriplectic Conditional Flow Matching for Dissipative Dynamics]{Metriplectic Conditional Flow Matching for Dissipative Dynamics}

\optauthor{
  \Name{Ali Baheri} \Email{akbeme@rit.edu}\\
  \addr Rochester Institute of Technology, Rochester, NY, USA
  \AND
  \Name{Lars Lindemann} \Email{llindemann@ethz.ch}\\
  \addr Automatic Control Laboratory, ETH Z\"urich, Switzerland
}

\begin{document}

\maketitle

\begin{abstract}
Metriplectic conditional flow matching (MCFM) learns dissipative dynamics without violating first principles. Neural surrogates often inject energy and destabilize long-horizon rollouts; MCFM instead builds the conservative--dissipative split into both the vector field and a structure-preserving sampler. MCFM trains via conditional flow matching on short transitions, avoiding long-rollout adjoints. In inference, a Strang-prox scheme alternates a symplectic update with a proximal metric step, ensuring discrete energy decay; an optional projection enforces strict decay when a trusted energy is available. We provide continuous- and discrete-time guarantees linking this parameterization and sampler to conservation, monotonic dissipation, and stable rollouts. On a controlled mechanical benchmark, MCFM yields phase portraits closer to ground truth and markedly fewer energy-increase and positive energy-rate events than an equally expressive unconstrained neural flow, while matching terminal distributional fit. 

\end{abstract}




\section{Introduction}


Learning dynamical models directly from data supports forecasting, control, and simulation in science and engineering \citep{Chen2018NODE}. In practice, however, purely neural surrogates often violate basic physical laws: they inject energy, drift off invariant manifolds, and deteriorate over long horizons, issues well-documented in geometric integration and structure-preserving modeling \citep{Hairer2006GNI}. These failures are most visible in \emph{dissipative} systems, where the correct behavior is a monotone loss of energy \citep{Morrison1986Metriplectic,Ottinger2005BET}. Most existing approaches either (i) impose soft penalties or physics-informed residuals that encourage energy decay \citep{Raissi2019PINN}, or (ii) adopt structures tailored to \emph{conservative} mechanics (e.g., Hamiltonian/Lagrangian networks and symplectic/variational integrators) \citep{Cranmer2020LNN,Zhong2020SymODEN,Saemundsson2020VIGN,Hairer2006GNI}. The former offer no guarantees and are sensitive to weights and step sizes; the latter excel on closed systems but do not address dissipation and can misrepresent frictional effects \citep{Morrison1986Metriplectic,Grmela1997GENERIC}. Recent diffusion/flow-based learners capture local velocities well from short transitions \citep{Ho2020DDPM,Song2021SBM,DeBortoli2021DSB,Lipman2023FlowMatching,Tong2023CFM,Albergo2023StochasticInterpolants}, yet unconstrained parameterizations likewise permit non-physical energy creation during rollout. The gap is a method that \emph{learns from data} while \emph{enforcing} the conservative--dissipative split that governs real-world dynamics \citep{Morrison1986Metriplectic,Grmela1997GENERIC}.

We address this gap with metriplectic conditional flow matching (MCFM). Rather than adding more regularizers, MCFM builds the physical law into the model and sampler. We represent the learned vector field as a metriplectic composition: a conservative channel that preserves a learned energy-like quantity and a dissipative channel that can only remove it. Training uses conditional flow matching on short, consecutive samples, keeping optimization simple and data-efficient. In inference, a structure-preserving splitting scheme alternates a symplectic update for the conservative part with a proximal ``shrink'' for the dissipative part, yielding discrete-time trajectories with near-monotone energy. The result is a drop-in replacement for unconstrained neural flows that produces stable and physically consistent rollouts without hand-tuned penalties.

\noindent \textbf{Contributions.}
(1) We propose \emph{metriplectic conditional flow matching}, a structure-first parameterization to learn the dissipative dynamics that encodes the conservative--dissipative split by design.  
(2) We pair the model with a structure-preserving sampler that guarantees discrete energy decay under mild conditions.  
(3) We provide theoretical guarantees linking the parameterization and sampler to conservation/monotonicity and rollout stability.

\section{Related Work}
\label{sec:related}

\noindent \textbf{Structure in learned dynamics.}
The vast literature embeds geometric and variational structures to improve long-horizon stability. Hamiltonian and Lagrangian neural networks preserve invariants of conservative mechanics \citep{Greydanus2019HNN,Cranmer2020LNN}, with symplectic and variational integrator-based discretizations further reducing drift \citep{Zhong2020SymODEN,Saemundsson2020VIGN,Hairer2006GNI}. Beyond conservative settings, physics-informed constraints \citep{Raissi2019PINN} and Koopman-based linearizations \citep{Lusch2018Koopman} aim at data efficiency, but lack a treatment of dissipation.

\noindent \textbf{Metriplectic and GENERIC formalisms.}
The metriplectic and GENERIC frameworks combine a skew (Hamiltonian) and a metric (dissipative) bracket to guarantee the conservation of an energy-like invariant and monotone decay of a dissipation potential \citep{Morrison1986Metriplectic,Grmela1997GENERIC,Ottinger2005BET}. These ideas have only recently begun to permeate machine learning; our contribution is to pair a learnable metriplectic parameterization with modern flow-based training and a structure-preserving sampler.

\noindent \textbf{Continuous-time models, diffusion, and flow matching.}
Neural ODEs provide continuous-time parameterizations for dynamics \citep{Chen2018NODE,Rubanova2019LatentODE,Kidger2020NeuralCDE}. Diffusion/score-based generative modeling \citep{Ho2020DDPM,Song2021SBM} and Schr\"odinger bridges \citep{DeBortoli2021DSB} motivate training via vector-field regression rather than simulation. Flow matching and conditional flow matching offer simple and simulation-free objectives to learn transport fields \citep{Lipman2023FlowMatching,Tong2023CFM,Albergo2023StochasticInterpolants}. Most prior approaches adopt unconstrained vector fields; instead, we impose a metriplectic decomposition and couple it with a Strang--prox integrator, yielding discrete-time energy monotonicity without penalty tuning.

\section{Methodology}
\label{sec:method}

We propose \emph{metriplectic conditional flow matching}, a learning framework for dissipative dynamics that embeds the fundamental split between conservative and dissipative effects directly into the vector field. Let $x\in\mathbb{R}^d$ denote the state and $t\in[0,1]$ a normalized time variable. We model trajectories as solutions of the probability-flow ODE
\begin{equation}
\dot{x} \;=\; v_\theta(x,t).
\end{equation}
We target finite-dimensional dissipative systems whose dynamics admit a metriplectic decomposition. The state $x \in \mathbb{R}^d$ may, in mechanical examples, split as $x=(q, p)$; explicit time-dependence enters via Fourier features $\phi(t)$. We assume $H_\theta, \Phi_\theta \in C^2$, a skew operator $J(x, t)^{\top}=-J(x, t)$, and a positive semidefinite metric $G_\theta(x, t) \succeq 0$. We enforce the standard degeneracy conditions

\begin{equation}
G_\theta \nabla H_\theta=0, \quad J \nabla \Phi_\theta=0
\end{equation}
either hard (by construction) or soft (small penalties), ensuring that the conservative channel cannot change $\Phi_\theta$ and the dissipative channel cannot change $H_\theta$.
We parameterize $v_\theta$ by the metriplectic decomposition
\begin{equation}
v_\theta(x,t)\;=\;J(x,t)\,\nabla H_\theta(x,t)\;-\;G_\theta(x,t)\,\nabla \Phi_\theta(x,t),
\label{eq:metriplectic}
\end{equation}
where $H_\theta$ is a learned invariant (e.g. energy), $\Phi_\theta$ is a learned dissipation potential (e.g., entropy or Lyapunov), $J$ is skew-symmetric $(J^\top=-J)$, and $G_\theta\succeq 0$ is positive semidefinite. We enforce the standard degeneracy conditions $G_\theta\nabla H_\theta=0$ and $J\nabla\Phi_\theta=0$, which ensure that the Hamiltonian part does not change $\Phi_\theta$ and the metric part does not change $H_\theta$. Consequently, along any solution $x(t)$,

\begin{equation}
\frac{d}{d t} H_\theta(x(t), t)=0, \quad \frac{d}{d t} \Phi_\theta(x(t), t)=-\left\|\nabla \Phi_\theta(x(t), t)\right\|_{G_\theta(x(t), t)}^2 \leq 0
\end{equation}
where $\|z\|_G^2:=z^{\top} G z$. The G-norm notation makes explicit that the dissipation rate is a non-negative quadratic form.

\noindent \textbf{Parameterization.} We represent $H_\theta$ and $\Phi_\theta$ with small multilayer perceptrons that take $[x,\phi(t)]$ as input, where $\phi(t)$ is a Fourier time embedding. Structure matrices are chosen to make the constraints trivial to satisfy: $J$ is fixed skew (e.g., canonical symplectic blocks) and $G_\theta=L_\theta L_\theta^\top$ with a low-rank factor $L_\theta$. Degeneracy is realized either \emph{hard}, by restricting the range of $G_\theta$ to a designated subspace or \emph{soft} by adding light penalties $\|G_\theta\nabla H_\theta\|^2+\|J\nabla\Phi_\theta\|^2$ to the objective. When a trusted physical invariant $E_{\mathrm{phys}}$ is known, we set $H_\theta\equiv E_{\mathrm{phys}}$ and only learn dissipation; otherwise, both $H_\theta$ and $\Phi_\theta$ are learned from the data.\footnote{\(E_{\mathrm{phys}}\) is the known physical energy/invariant. 
For mechanical states \(x=(q,p)\) with mass matrix \(M\) and potential \(V(q)\),
\(E_{\mathrm{phys}}(q,p)=\tfrac12 p^\top M^{-1}p+V(q)\), with 
\(\nabla E_{\mathrm{phys}}=[\,\nabla_q V(q);\; M^{-1}p\,]\).
If no trusted invariant is available, we learn \(H_\theta\) instead of fixing \(H_\theta\equiv E_{\mathrm{phys}}\).}

\noindent \textbf{Learning with conditional flow matching.} Training uses only short consecutive samples $(x_k,x_{k+1})$ collected at step $\Delta t$. For each pair, we sample $\tau\sim\mathrm{Unif}[0,1]$, form the linear bridge $x_\tau=(1-\tau)x_k+\tau x_{k+1}$, and define the conditional target velocity.
\begin{equation}
u_\tau(x_\tau)=\frac{x_{k+1}-x_k}{\Delta t}.
\end{equation}
The parameters are obtained by minimizing the regression loss
\begin{equation}
\mathcal{L}_{\mathrm{CFM}}(\theta)=
\mathbb{E}_{(k,\tau)} \big\|v_\theta(x_\tau,\tau)-u_\tau(x_\tau)\big\|^2
\;+\;\lambda_{\mathrm{soft}}\!\Big(\|G_\theta\nabla H_\theta\|^2+\|J\nabla\Phi_\theta\|^2\Big)
\;+\;\lambda_{\mathrm{reg}}\!\Big(\|\nabla H_\theta\|_2^2+\|G_\theta\|_F^2\Big),
\end{equation}
optimized with AdamW and cosine annealing; gradients are clipped. Unlike adjoint-based training through long rollouts, conditional flow matching keeps optimization simple and stable.

\noindent \textbf{Physics-aware sampling.} In inference, we integrate $\dot x=v_\theta(x,t)$ on a finite horizon $T$. To preserve the metriplectic structure discretely, we use a Strang splitting scheme: a Hamiltonian half-step on $\dot x=J\nabla H_\theta$ using a symplectic update (e.g. velocity-Verlet), a metric step on $\dot x=-G_\theta\nabla\Phi_\theta$ (which, for common parameterizations, has a closed-form proximal ``shrink'' in the dissipative coordinates), and a second Hamiltonian half-step. This sampler inherits the continuous guarantees---phase-space volume is preserved by the conservative sub-step while the dissipation potential decreases monotonically during the metric sub-step. When strict monotonicity of a known physical energy is required, we append a light projection
\begin{equation}
v \;\leftarrow\; v \;-\; \frac{\max\{0,\langle\nabla E_{\mathrm{phys}},v\rangle\}}{\|\nabla E_{\mathrm{phys}}\|^2+\varepsilon}\,\nabla E_{\mathrm{phys}},
\end{equation}
which enforces $\nabla E_{\mathrm{phys}}\!\cdot v\le 0$ without altering the tangential motion.

\noindent \textbf{Scope and comparison.}
The unstructured baseline used in the experiments is a standard neural velocity field $v_\theta(x,t)=f_\theta([x,\phi(t)])$ trained with the same conditional flow matching objective and integrated with a high-order solver. The baseline captures local trends but lacks invariance and monotonicity guarantees, which manifest as energy drift or creation over long horizons. In contrast, MCFM restricts the hypothesis class to metriplectic fields and pairs it with a matching sampler, yielding trajectories that are faithful to both the observed data and the governing physical principles.

\section{Main theoretical results}
We recall the metriplectic parameterization in Eq.~(3):
\[
\dot x \;=\; v_\theta(x,t)\;=\;J(x,t)\nabla H_\theta(x,t)\;-\;G_\theta(x,t)\nabla \Phi_\theta(x,t),
\]
with $J^\top=-J$, $G_\theta\succeq 0$, and the degeneracy conditions $G_\theta\nabla H_\theta=0$ and $J\nabla \Phi_\theta=0$.
Unless noted otherwise, $H_\theta,\Phi_\theta\in C^2$ are time-independent, and the Hamiltonian sub-integrator used in the sampler is symplectic and time-reversible (order $p\!\ge\!2$).

\begin{theorem}[Continuous-time conservation and dissipation]
Along any solution of $\dot x=v_\theta$ with the degeneracy conditions of Eq.~(2) enforced,
\[
\frac{d}{dt}H_\theta(x(t))=0
\qquad\text{and}\qquad
\frac{d}{dt}\Phi_\theta(x(t))=-\,\nabla\Phi_\theta(x(t))^\top G_\theta(x(t))\,\nabla\Phi_\theta(x(t))\;\le 0.
\]
\emph{Proof sketch.} By the chain rule,
$\dot H_\theta=\nabla H_\theta^\top J\nabla H_\theta-\nabla H_\theta^\top G_\theta\nabla \Phi_\theta=0-0$
(using $J^\top=-J$ and $G_\theta\nabla H_\theta=0$).
Similarly,
$\dot \Phi_\theta=\nabla\Phi_\theta^\top J\nabla H_\theta-\nabla\Phi_\theta^\top G_\theta\nabla\Phi_\theta
=0-\nabla\Phi_\theta^\top G_\theta\nabla\Phi_\theta\le 0$
(using $J\nabla\Phi_\theta=0$ and $G_\theta\succeq 0$).
\end{theorem}

\begin{corollary}[Strict decay and attractivity]
If $G_\theta(x)\succeq \mu I$ on its range for some $\mu>0$, then $\dot\Phi_\theta<0$ whenever $\nabla\Phi_\theta\neq 0$.
Moreover, the largest invariant set contained in $\{\nabla\Phi_\theta=0\}$ is Lyapunov attractive (LaSalle).
\end{corollary}
\setcounter{theorem}{1}
\begin{theorem}[Discrete-time guarantees for the Strang--prox sampler]
Consider one Strang step with stepsize $h>0$: (i) Hamiltonian half-step under $\dot x=J\nabla H_\theta$ (symplectic, order $p\!\ge\!2$), (ii) metric step under $\dot x=-G_\theta\nabla\Phi_\theta$ (prox/implicit-Euler in dissipative coordinates), (iii) Hamiltonian half-step.

\emph{(i) Near-conservation of $H_\theta$.}
Because the metric flow cannot change $H_\theta$ by degeneracy, only the symplectic substeps contribute.
Time-reversible symplectic schemes preserve a modified Hamiltonian $\tilde H_\theta=H_\theta+O(h^p)$, hence over a full Strang step
\[
\bigl|H_\theta(x_{k+1})-H_\theta(x_k)\bigr|\;=\;O(h^{p+1}).
\]

\emph{(ii) Monotone decrease of $\Phi_\theta$ up to $O(h^2)$.}
The Hamiltonian substeps leave $\Phi_\theta$ unchanged (by $J\nabla\Phi_\theta=0$).
The metric substep is a gradient flow in the $G_\theta$-metric; its proximal discretization yields
\[
\Phi_\theta(x_{k+1})-\Phi_\theta(x_k)\;\le\; -\,h\,\nabla\Phi_\theta(x_k)^\top G_\theta(x_k)\nabla\Phi_\theta(x_k)\;+\;O(h^2).
\]
If $G_\theta\succeq \mu I$ on its range and $h$ is sufficiently small (e.g., $h\!\le\!\mu/L_\Phi$ for an $L_\Phi$-smooth $\nabla\Phi_\theta$), the right side is strictly negative whenever $\nabla\Phi_\theta(x_k)\neq 0$.
\end{theorem}


\section{Empirical Study}
\label{sec:empirical}

\noindent \textbf{Task and data.}
We evaluated MCFM on a controlled dissipative system: the damped pendulum. Training pairs are short one-step segments obtained from the ground-truth ODE $\dot q=p/M,\;\dot p=-MgL\sin q-\gamma p$ with $\gamma$ sampled uniformly in $[0.05,0.30]$, step $\Delta t=0.1$, and $500$ trajectories for training and $100$ for testing. This setting isolates the question of \emph{physical consistency under dissipation} while keeping the supervision simple (short transitions rather than long rollouts).

\begin{figure}[t]
  \centering
  \includegraphics[width=0.5\textwidth]{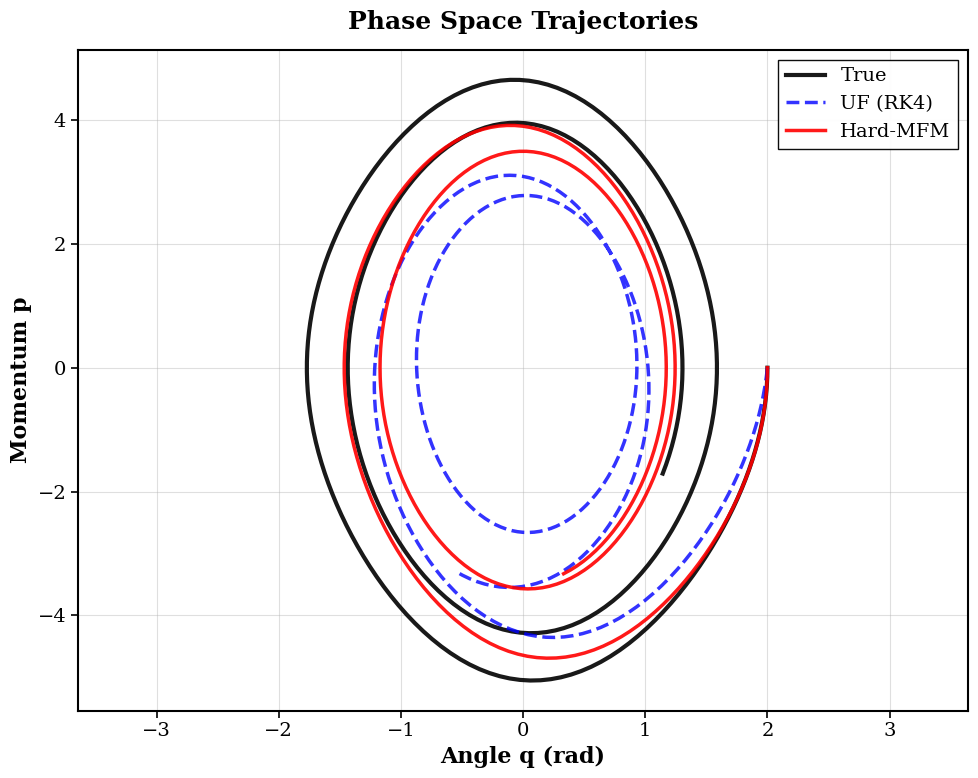}
  \caption{Phase space trajectories (True, UF, Hard--MCFM).}
  \label{fig:phase}
\end{figure}

\noindent \textbf{Models and training.}
The baseline (\textbf{UF}) is an unconstrained neural velocity field $v_\theta(x,t)$ with Fourier time features; it is trained by conditional flow matching (CFM) on the same pairs as MCFM. Our \textbf{Hard--MCFM} parameterizes $v_\theta$ through a metriplectic split $J\nabla H_\theta - G_\theta\nabla\Phi_\theta$ with fixed skew $J$ and learned $G_\theta\succeq 0$, enforcing the conservative--dissipative decomposition by design. At test time, UF is integrated by RK4; MCFM uses a Strang splitting sampler (symplectic Hamiltonian half-steps plus a proximal metric \say{shrink} $p\!\leftarrow\!p/(1+\gamma_\theta\Delta t)$) to preserve the structure discretely. 

\begin{figure}[t]
  \centering
  \includegraphics[width=0.85\textwidth]{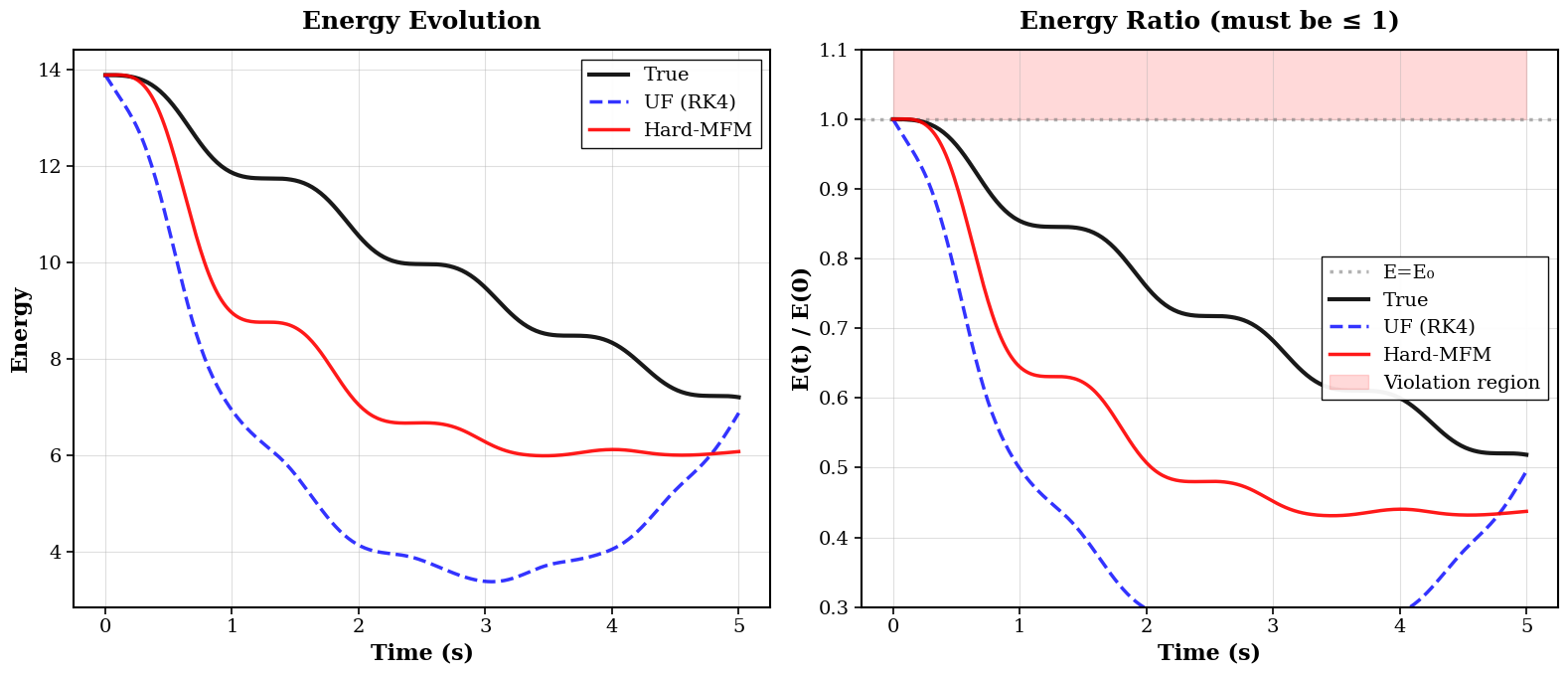}
  \caption{Energy evolution (left) and energy ratio (right). Violation region shaded.}
  \label{fig:energy}
\end{figure}

\noindent \textbf{Metrics.}
Over a $5$\,s rollout we report: (i) energy \emph{increase fraction} $\tfrac{1}{T}\sum_t\mathbf{1}[E_{t+1}>E_t]$ (target $0$); (ii) \emph{max energy ratio} $\max_t E_t/E_0$ (target $\le1$); (iii) \emph{final energy ratio} $E_T/E_0$ (smaller is more dissipative); (iv) \emph{sign-error fraction} $\tfrac{1}{T}\sum_t\mathbf{1}[\dot E(t)>0]$; and (v) a terminal-distribution fit via sliced Wasserstein-2 (SW-2). Together, these capture geometry, monotonicity, and distributional accuracy.

\noindent \textbf{Results.}
Figure~\ref{fig:phase} shows that UF rapidly collapses toward the origin and drifts from the outer level sets, while MCFM tracks the geometry of the true flow for the full horizon. This reflects the metriplectic bias: Hamiltonian sub-steps move tangentially on learned level sets, and dissipation acts only along the dissipative coordinates. In Figure~\ref{fig:energy}, all models initially dissipate; however, UF exhibits late-horizon energy \emph{increases} (the ratio $E(t)/E(0)$ rebounds above its earlier minimum), while MCFM maintains a smooth, largely monotone decay with small oscillations. Early-time over-damping is visible for both models but is less severe for MCFM relative to the reference curve.

\begin{figure}[t]
  \centering
  \begin{minipage}{0.45\textwidth}
    \centering
    \includegraphics[width=\textwidth]{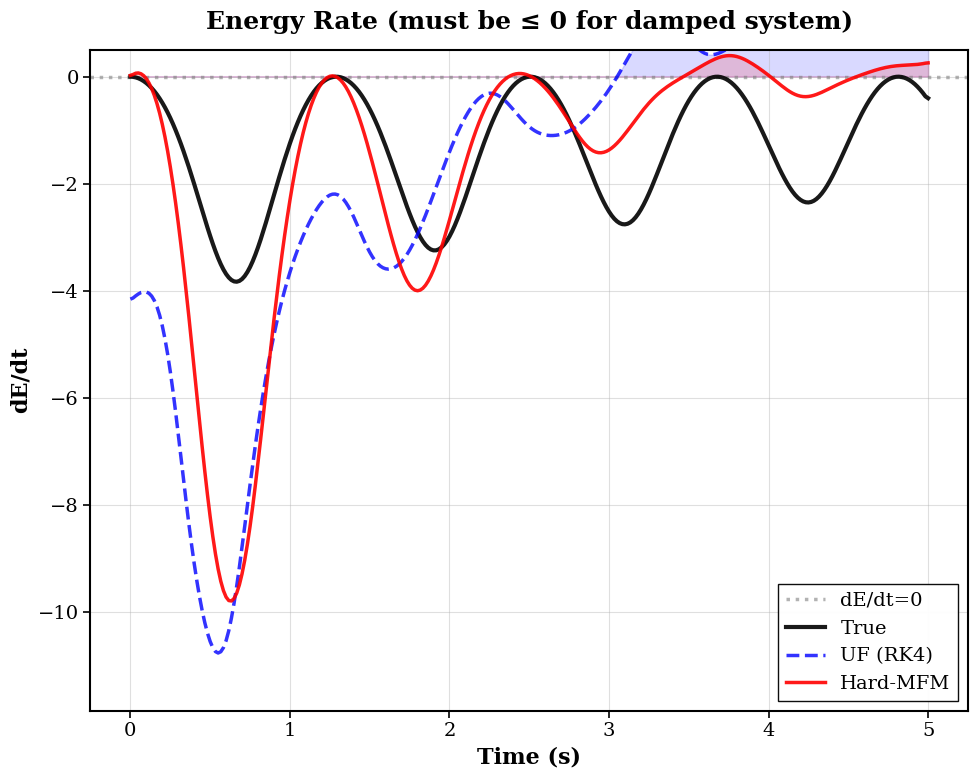}
    \caption{Energy rate $dE/dt$ (must be $\le 0$).}
    \label{fig:edot}
  \end{minipage}\hfill
  \begin{minipage}{0.50\textwidth}
    \centering
    \includegraphics[width=\textwidth]{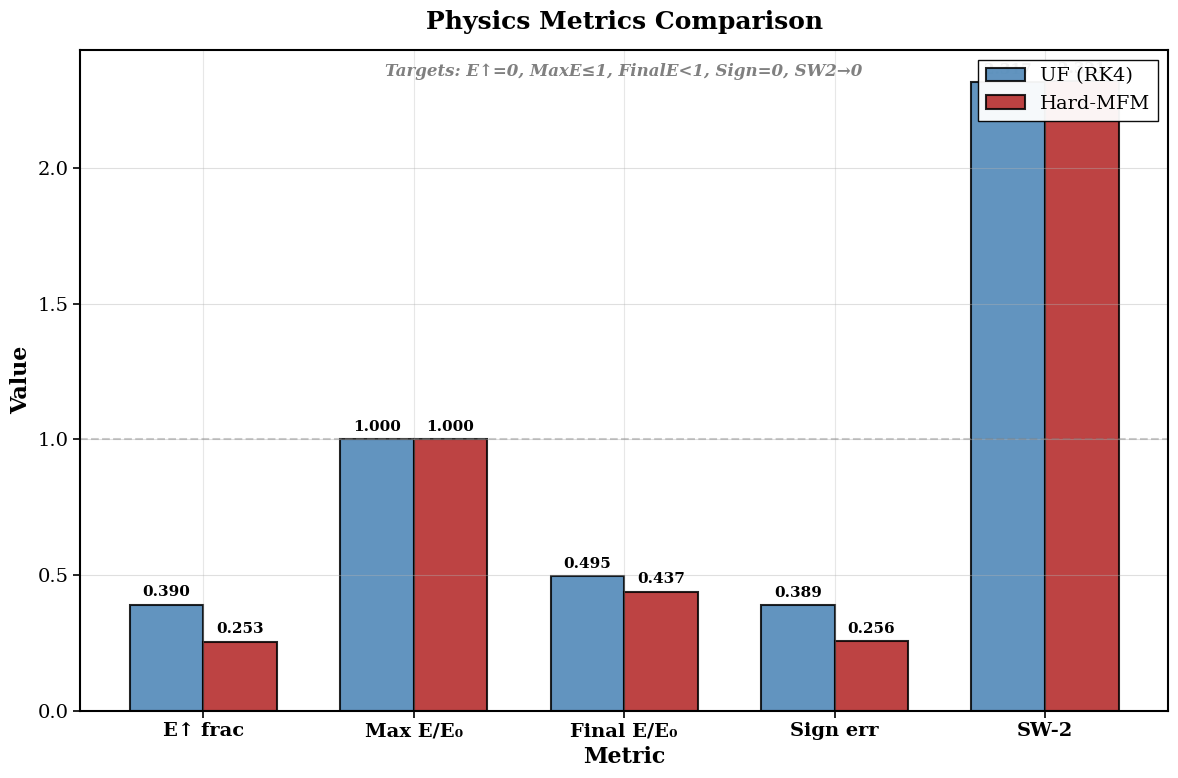}
    \caption{Physics metrics comparison (lower is better except SW-2).}
    \label{fig:metrics}
  \end{minipage}
\end{figure}

Figure~\ref{fig:edot} plots $dE/dt$. The shaded region indicates forbidden positive rates for a damped system. UF spends substantial late-horizon time with $\dot E>0$; MCFM's violations are sparse and of small magnitude, staying near the zero line, consistent with the proximal metric step and our projection that removes any residual positive component along $\nabla E$. The bar graph in Figure~\ref{fig:metrics} summarizes the above. MCFM reduces the energy-increase fraction and sign-error fraction relative to UF, while matching terminal-distribution quality (SW-2). In short, MCFM achieves \emph{stronger physical consistency} without sacrificing distributional fit.

\section{Conclusions}

We proposed metriplectic conditional flow matching, a deterministic probability-flow framework that embeds conservative--dissipative structure via a metriplectic split, with hard/soft degeneracy and optional use of a known energy. Trained by conditional flow matching and paired with a physics-aware sampler (Strang splitting plus an optional energy-monotonicity projection), MCFM aims to produce stable, physically consistent rollouts without expensive long-horizon training. Initial demonstrations on canonical dissipative systems illustrate its practicality. Future directions include richer skew/metric parameterizations, extension to PDEs and control, and tighter theoretical guarantees on long-horizon behavior.

\newpage

\bibliography{sample}

\newpage
\appendix
\section{Proof of Theorem~1 (Continuous-time conservation and dissipation)}
We recall the parametrization in Eq.~(2):
\[
\dot x \;=\; v_\theta(x,t)\;=\;J(x,t)\nabla H_\theta(x,t)\;-\;G_\theta(x,t)\nabla \Phi_\theta(x,t),
\]
where $J^\top=-J$, $G_\theta\succeq 0$, and the degeneracy conditions
$G_\theta\nabla H_\theta=0$ and $J\nabla \Phi_\theta=0$ hold.
Throughout this proof $H_\theta,\Phi_\theta$ are $C^2$ and time-independent, and $x(\cdot)$ is a classical solution of $\dot x=v_\theta$.

\begin{proof}[Proof of Theorem~1]
Set $H:=H_\theta$, $\Phi:=\Phi_\theta$, $J:=J(\cdot)$, $G:=G_\theta(\cdot)$ for brevity.

\paragraph{Step 1: Conservation of $H$.}
By the chain rule along $x(t)$,
\[
\frac{d}{dt}H(x(t))
=\nabla H(x(t))^\top \dot x(t)
=\nabla H^\top\!\bigl(J\nabla H\bigr)\;-\;\nabla H^\top\!\bigl(G\nabla \Phi\bigr).
\]
The first term vanishes because $J$ is skew-symmetric:
for any $a\in\mathbb{R}^d$, $a^\top J a=(a^\top J a)^\top=a^\top J^\top a=-a^\top J a$, hence $a^\top J a=0$.
The second term vanishes by the degeneracy condition $G\nabla H=0$.
Therefore $\tfrac{d}{dt}H(x(t))=0$ for all $t$, i.e., $H(x(t))\equiv H(x(0))$.

\paragraph{Step 2: Dissipation of $\Phi$.}
Again by the chain rule,
\[
\frac{d}{dt}\Phi(x(t))
=\nabla \Phi(x(t))^\top \dot x(t)
=\nabla \Phi^\top\!\bigl(J\nabla H\bigr)\;-\;\nabla \Phi^\top\!\bigl(G\nabla \Phi\bigr).
\]
The first term is zero by $J\nabla \Phi=0$ (degeneracy).
For the second term, positive semidefiniteness of $G$ yields
$\nabla \Phi^\top G\nabla \Phi \ge 0$,
so
\[
\frac{d}{dt}\Phi(x(t)) \;=\; -\,\nabla \Phi(x(t))^\top G(x(t))\,\nabla \Phi(x(t)) \;\le\;0.
\]
Integrating in time gives the exact energy--dissipation identity
\[
\Phi\bigl(x(T)\bigr)-\Phi\bigl(x(0)\bigr)
= -\int_0^T \nabla \Phi\bigl(x(t)\bigr)^\top G\bigl(x(t)\bigr)\,\nabla \Phi\bigl(x(t)\bigr)\,dt \;\le\; 0,
\]
so $\Phi(x(t))$ is non-increasing along trajectories.

\paragraph{Conclusion.}
Combining Steps 1--2 proves that along any solution of Eq.~(2) with the stated degeneracy conditions,
$\tfrac{d}{dt}H(x(t))=0$ and $\tfrac{d}{dt}\Phi(x(t))=-\nabla \Phi^\top G\nabla \Phi\le 0$.
\end{proof}

\paragraph{Remarks.}
(i) If $G(x)$ is positive definite in its action subspace (there exists $\mu>0$ with $\xi^\top G(x)\xi\ge \mu\|\xi\|^2$ for all $\xi$ in that subspace), then $\tfrac{d}{dt}\Phi(x(t))<0$ whenever $\nabla\Phi(x(t))\neq 0$ (used in the corollary in the main text).
(ii) If $H$ or $\Phi$ depend explicitly on $t$, the same computation yields
$\tfrac{d}{dt}H(x(t),t)=\partial_t H(x(t),t)$ and
$\tfrac{d}{dt}\Phi(x(t),t)=\partial_t \Phi(x(t),t)-\nabla \Phi^\top G\nabla \Phi$.

\newpage

\section{Proof of Theorem~2 (Discrete-time guarantees for the Strang--prox sampler)}
Recall the parameterization in Eq.~(2):
\[
\dot x \;=\; J(x,t)\nabla H_\theta(x,t)\;-\;G_\theta(x,t)\nabla \Phi_\theta(x,t),
\]
with $J^\top=-J$, $G_\theta\succeq 0$, and degeneracy $G_\theta\nabla H_\theta=0$, $J\nabla\Phi_\theta=0$.
We analyze one Strang step with step size $h>0$:
\[
x_{k+\frac12}=\Psi^{\mathrm H}_{h/2}(x_k),\quad
x_{k+\frac12}^-=\Psi^{\mathrm D}_{h}(x_{k+\frac12}),\quad
x_{k+1}=\Psi^{\mathrm H}_{h/2}(x_{k+\frac12}^-),
\]
where $\Psi^{\mathrm H}_{\tau}$ is a symplectic, time-reversible order-$p\!\ge\!2$ integrator for $\dot x=J\nabla H_\theta$ and $\Psi^{\mathrm D}_{\tau}$ is the metric/prox step for $\dot x=-G_\theta\nabla \Phi_\theta$ (implemented in dissipative coordinates so that the degeneracy w.r.t.\ $H_\theta$ is enforced within the substep).
Throughout we assume $H_\theta,\Phi_\theta\in C^2$, $\nabla\Phi_\theta$ is $L_\Phi$-Lipschitz in the compact region visited, and $G_\theta$ is locally Lipschitz and uniformly bounded on that region. For brevity, write $H,\Phi,G$.

\subsection*{(i) Near-conservation of $H$}
\paragraph{Step A (metric substep).}
By degeneracy, $\nabla H^\top (G\nabla\Phi)=0$ pointwise. Since the metric subflow follows directions in $\mathrm{range}(G)$ and the implementation enforces the degeneracy within the substep, the metric map leaves $H$ unchanged:
\[
H\bigl(\Psi^{\mathrm D}_{h}(x)\bigr)=H(x).
\]

\paragraph{Step B (Hamiltonian half-steps).}
Let $\Psi^{\mathrm H}_{\tau}$ be a symmetric symplectic method of (even) order $p\ge 2$ for $\dot x=J\nabla H$. Classical backward error analysis yields a \emph{modified Hamiltonian} $\tilde H_\tau=H+O(\tau^p)$ such that one step of size $\tau$ preserves $\tilde H_\tau$ exactly and the local error in $H$ is $O(\tau^{p+1})$.
Hence for each half-step of length $\tau=h/2$,
\[
\bigl|H(x_{k+\frac12})-H(x_k)\bigr|=O\!\left((h/2)^{p+1}\right),\qquad
\bigl|H(x_{k+1})-H(x_{k+\frac12}^-)\bigr|=O\!\left((h/2)^{p+1}\right).
\]

\paragraph{Conclusion for (i).}
Combining Steps A--B and using the triangle inequality,
\[
\bigl|H(x_{k+1})-H(x_k)\bigr|
\;\le\; O\!\left((h/2)^{p+1}\right) + \underbrace{\bigl|H(x_{k+\frac12}^-)-H(x_{k+\frac12})\bigr|}_{=\,0} + O\!\left((h/2)^{p+1}\right)
\;=\; O(h^{p+1}).
\]

\subsection*{(ii) Monotone decrease of $\Phi$ up to $O(h^2)$}
\paragraph{Step A (Hamiltonian half-steps).}
By degeneracy $J\nabla\Phi=0$, the Hamiltonian vector field is tangent to $\{\Phi=\text{const}\}$. Thus, the exact Hamiltonian flow leaves $\Phi$ invariant, and a symmetric order-$p$ method changes $\Phi$ at most by $O((h/2)^{p+1})$ per half-step. Since $p\ge2$, these contributions are $O(h^3)$ and can be absorbed in $O(h^2)$ below.

\paragraph{Step B (metric substep).}
Consider the metric ODE $\dot x=-G(x)\nabla\Phi(x)$ and one (prox/implicit-Euler) step
\[
x^-=x - h\,G(x^-)\nabla\Phi(x^-).
\]
The standard descent lemma for $L_\Phi$-smooth $\Phi$ gives
\[
\Phi(x^-)\;\le\;\Phi(x)\;+\;\nabla\Phi(x)^\top(x^-\!-\!x)\;+\;\tfrac{L_\Phi}{2}\|x^-\!-\!x\|^2.
\]
Using the implicit update and local Lipschitzness of $G$,
\[
x^-\!-\!x \;=\; -h\,G(x)\nabla\Phi(x) \;+\; O(h^2),
\]
whence
\[
\Phi(x^-)-\Phi(x)
\;\le\; -\,h\,\nabla\Phi(x)^\top G(x)\nabla\Phi(x)\;+\;O(h^2).
\]

\paragraph{Conclusion for (ii).}
In one full Strang step,
\[
\Phi(x_{k+1})-\Phi(x_k)
\;=\; \underbrace{\bigl[\Phi(x_{k+\frac12})-\Phi(x_k)\bigr]}_{O(h^3)}
\;+\;\underbrace{\bigl[\Phi(x_{k+\frac12}^-)-\Phi(x_{k+\frac12})\bigr]}_{\le -\,h\,\nabla\Phi^\top G \nabla\Phi\,+\,O(h^2)}
\;+\;\underbrace{\bigl[\Phi(x_{k+1})-\Phi(x_{k+\frac12}^-)\bigr]}_{O(h^3)},
\]
which yields
\[
\Phi(x_{k+1})-\Phi(x_k)\;\le\;-\,h\,\nabla\Phi(x_k)^\top G(x_k)\nabla\Phi(x_k)\;+\;O(h^2).
\]
If in addition $G(x)\succeq \mu I$ on its range and $h$ is sufficiently small, the $O(h^2)$ term is dominated and the right-hand side is strictly negative whenever $\nabla\Phi(x_k)\neq 0$.
This proves Theorem~2.

\paragraph{Remark (explicit stepsize condition).}
If $G(x)\succeq \mu I$ on its range and $\|G(x)\|\le M$ on the region of interest, then
\[
-\,h\,\nabla\Phi^\top G\nabla\Phi \;\le\; -\,h\,\mu\,\|\nabla\Phi\|^2,
\qquad
O(h^2)\;\le\; C\,h^2\,\|\nabla\Phi\|^2
\]
for some $C$ depending on $L_\Phi$ and local Lipschitz bounds of $G$.
Thus any $h\le \mu/C$ ensures $\Phi(x_{k+1})<\Phi(x_k)$ whenever $\nabla\Phi(x_k)\neq 0$.

\section{Proof of Corollary~2 (``Hard'' monotonicity with projection)}
Assume a trusted energy $E_{\mathrm{phys}}$ and the projection in Eq.~(6), which enforces
$\nabla E_{\mathrm{phys}}(x)^\top v_{\mathrm{proj}}(x)\le 0$ while leaving tangential components unchanged.
Let $E:=E_{\mathrm{phys}}$ be $L$-smooth and consider the explicit step $x^+=x+h\,v_{\mathrm{proj}}(x)$.
By the descent lemma for $L$-smooth functions,
\[
E(x^+)\;\le\; E(x)\;+\;h\,\nabla E(x)^\top v_{\mathrm{proj}}(x)\;+\;\tfrac{L}{2}h^2\|v_{\mathrm{proj}}(x)\|^2.
\]
Because $\nabla E^\top v_{\mathrm{proj}}\le 0$ by construction, we have
\[
E(x^+)\;\le\;E(x)\;-\;h\,\bigl[\nabla E(x)^\top v(x)\bigr]_+\;+\;\tfrac{L}{2}h^2\|v_{\mathrm{proj}}(x)\|^2.
\]
Hence $E(x^+)\le E(x)$ for all sufficiently small $h$, and $E$ decreases strictly whenever $\bigl[\nabla E(x)^\top v(x)\bigr]_+>0$.

\end{document}